\documentclass[sigconf]{acmart}

\AtBeginDocument{%
  \providecommand\BibTeX{{%
 \normalfont B\kern-0.5em{\scshape i\kern-0.25em b}\kern-0.8em\TeX}}}

\copyrightyear{2022}
\acmYear{2022}
\setcopyright{acmcopyright}\acmConference[WWW '22]{Proceedings of the ACM Web Conference 2022}{April 25--29, 2022}{Virtual Event, Lyon, France}
\acmBooktitle{Proceedings of the ACM Web Conference 2022 (WWW '22), April 25--29, 2022, Virtual Event, Lyon, France}
\acmPrice{15.00}
\acmDOI{10.1145/3485447.3511927}
\acmISBN{978-1-4503-9096-5/22/04}

\acmSubmissionID{328}

\usepackage{url}
\usepackage{amsmath}
\usepackage{algorithm}
\usepackage{algorithmic}
\usepackage{epstopdf}
\usepackage{appendix}
\usepackage{multirow}
\usepackage{makecell}
\usepackage{graphicx}
\usepackage{enumitem}
\usepackage{bm}
\usepackage{soul,color}
\soulregister\cite7
\soulregister\ref7
\soulregister\pageref7
\begin{document}

\title{Swift and Sure: Hardness-aware Contrastive Learning for Low-dimensional Knowledge Graph Embeddings}

\author{Kai Wang$^{1,2}$*, Yu Liu$^{1}$, Quan Z. Sheng$^{2}$}
\thanks{*Corresponding author}
\affiliation{\textsuperscript{1}Key Laboratory for Ubiquitous Network and Service Software of Liaoning Province, School of Software, \\Dalian University of Technology \city{Dalian} \state{Liaoning} \country{China} \postcode{116620}}
\affiliation{\textsuperscript{2}Intelligent Computing Laboratory, School of Computing, Macquarie University \city{Sydney} \state{NSW} \country{Australia} \postcode{2109}}
\email{kai_wang@mail.dlut.edu.cn, yuliu@dlut.edu.cn, michael.sheng@mq.edu.au}

\renewcommand{\authors}{Kai Wang, Yu Liu, and Quan Z. Sheng}
\renewcommand{\shortauthors}{K. Wang, Y. Liu, and Q. Z. Sheng}

\begin{abstract}
Knowledge graph embedding (KGE) has shown great potential in automatic knowledge graph (KG) completion and knowledge-driven tasks. However, recent KGE models suffer from high training cost and large storage space, thus limiting their practicality in real-world applications. To address this challenge, based on the latest findings in the field of Contrastive Learning, we propose a novel KGE training framework called Hardness-aware Low-dimensional Embedding (HaLE). Instead of the traditional Negative Sampling, we design a new loss function based on query sampling that can balance two important training targets, {\em Alignment} and {\em Uniformity}. Furthermore, we analyze the hardness-aware ability of recent low-dimensional hyperbolic models and propose a lightweight hardness-aware activation mechanism.
The experimental results show that in the limited training time, HaLE can effectively improve the performance and training speed of KGE models on five commonly-used datasets. After training just a few minutes, the HaLE-trained models are competitive compared to the state-of-the-art models in both low- and high-dimensional conditions.
\end{abstract}



\maketitle

\section{Introduction}
Knowledge Graph (KG) has shown great potential for recording semantic data from the Web as factual triples in the form of (head entity, relation, tail entity).
Knowledge Graph Embedding (KGE) can further represent entities and relations in the continuous vector space, and has been widely utilized in automatic KG completion and knowledge-driven tasks, such as information retrieval, semantic matching, and question answering \cite{DogICLR20,WWW21p1,WWW21p2,WWW21MixedCurv}.
However, recent KGE models usually utilize complicated computational structures and high-dimensional vectors up to 500 or even 1,000 dimensions \cite{RotatE,ConvE,ConvKB}.
Training such high-dimensional models demands prohibitive training costs and storage space, yet only achieving slight performance increase.
Meanwhile, large-scale KGs in real-world industrial applications, usually contain millions or billions of entities and need to be updated constantly based on real-time business data \cite{PCICLR20,PCKGEMNLP2020}.
Consequently, current KGE models mostly still stay in laboratory environments and remain difficult to be deployed in practical applications \cite{KGCompress, OurMulDE}.

To reduce the training costs, a promising way is to design new loss functions.
As the Negative Sampling loss has been indicated time-consuming and unstable, Sun et al. \cite{RotatE} proposed a Self-adversarial Negative Sampling loss,
which uses the Softmax-normalized triple score as the weight of each negative sample to accelerate model convergence.
Besides, recent research efforts from the community have
proposed several non-sampling training strategies in All-Negative or Non-Negative ways \cite{HEKGE-WWW21, HEKGE-NAACL21}.
Unfortunately, these methods still have respective constraints
and can only be applied to some specific KGE models.
To reduce the space complexity, low-dimensional hyperbolic-based KGE models have drawn attention, such as MuRP, RotH, RefH and AttH \cite{MuRP, GoogleAttH}.
Although they can achieve good performance when using 32 or 64 dimensions, the calculations in the hyperbolic space are much more complicated than those in the Euclidean space \cite{OurRotL}.

Two valuable insights from the latest Contrastive Learning studies inspire us to analyze previous KGE efforts from a new perspective.
First, Wang and Isola \cite{CL1-ICML20} identified key properties of Contrastive Learning, namely \textit{Alignment} and \textit{Uniformity}.
Training negative samples to form a uniform vector distribution is beneficial for learning separable features.
According to our analysis, different KGE training strategies
share the same goal.
Second, Wang and Liu \cite{CL2-CVPR21} discovered that the temperature $\tau$ in the contrastive loss has a hardness-aware capability to control the strength of penalties on different negative samples.
Similar attempts have been made in the hyperbolic geometry of recent hyperbolic KGE models and the self-adversarial loss \cite{RotatE}.
These insights motivate us to re-think the training target of KGE models and the weight assignment for different samples.

In this paper,
we propose a novel KGE training framework, \textbf{H}ardness-\textbf{a}ware \textbf{L}ow-dimensional \textbf{E}mbedding (\textbf{HaLE}).
To the best of our knowledge, our work is the first to achieve lightweight KGE training from a Contrastive Learning perspective.
Instead of simply applying Contrastive Learning methods,
we propose two innovative techniques: the {\em Query Sampling Loss} and the {\em Hardness-aware Activation Mechanism}.
To balance the two properties, Alignment and Uniformity, the new loss function maximizes the scores of all positive instances, while forcing all entity vectors to stay away from a limited number of sampled query vectors in the vector space.
As the result, the Query Sampling loss can provide stable gradients with small training costs.
Furthermore, we propose the Hardness-aware Activation mechanism with novel activation functions, $Hanon(\cdot)$ and $Halin(\cdot)$, to flexibly control the strength of penalties of easy and difficult instances.
Requiring few calculations, this mechanism has the potential to outperform the temperature trick and the hyperbolic geometry in hyperbolic-based KGE models.

We conduct extensive experiments on five commonly-used datasets, including FB15k-237, WN18RR, CoDEx-S/M/L.
The results show that HaLE can significantly improve the training speed of multiple KGE models.
Compared with previous KGE training strategies, the HaLE-trained models can obtain a higher prediction accuracy after training several minutes. Their performance is competitive compared to the state-of-the-art KGE models in both low- and high-dimensional conditions.

The rest of the paper is organized as follows.
We introduce the background and notations in Section \ref{sec:2}. Section \ref{sec:3} details the HaLE framework and its two major components.
Section \ref{sec:4} reports the experimental studies, and Section \ref{sec:5} further discusses the related work.
Finally, we provide some concluding remarks in Section \ref{sec:6}.

\vspace{-2mm}
\section{Background}
\label{sec:2}
In this section, we will briefly describe the Knowledge Graph Embedding and the Contrastive Learning techniques.
The main notations that will be used in this paper are summarized in Table~\ref{tab:1} in the Appendix~\ref{app:1}.

\vspace{-2mm}
\subsection{Knowledge Graph Embedding}
A Knowledge Graph $\mathcal{G}$ is composed by a collection of triples $(e_h,r,e_t)$, in which $r \in R$ is the relation between the head and tail entities $e_h,e_t \in E$.
A KGE model is usually trained by the link prediction task to represent each entity $e \in E$ (or relation $r \in R$) as a $d$-dimensional continuous vector.
Given a query $q=(e, r)$, link prediction aims to find the target entity $e_p \in E$ satisfying that $(e, r, e_p)$ or $(e_p, r, e)$ belongs to the knowledge graph $\mathcal{G}$.

To achieve this goal, the KGE model needs to score all candidate triples via a scoring function.
Given a triple $(e_h,r,e_t)$, the mainstream scoring function can be defined as $f(e_h,r,e_t)=\Psi(\Phi(e_h,r), e_t)$, which involves the following two operations:
\begin{itemize}
\item {\em Transform function} $\Phi(e_h,r)$ transforms the head vector $\mathbf{e}_h$ using the relation vector $\mathbf{r}$ and outputs the query vector $\mathbf{q}$;

\item {\em Similarity function} $\Psi(\mathbf{q},e_t)$ measures the similarity between the tail vector $\mathbf{e}_t$ and the transformed head vector $\mathbf{q}$.
\end{itemize}

Taken two typical KGE models, TransE \cite{TransE} and DistMult \cite{DistMult}, as examples, TransE's transform function is $\Phi(e_h,r)=\mathbf{e}_h+\mathbf{r}$ and its similarity function is equivalent to the $L_1$ or $L_2$ distance, while the transform function of DistMult is $\Phi(e_h,r)=\mathbf{e}_h \cdot \mathbf{r}$ and its similarity function is the dot product operation.
The similarity score $s$ produced by the scoring function $f(\cdot)$ is regarded as the triple score.
Most KGE models are trained by minimizing a Negative Sampling loss, to make the score of the qualified triple higher than those of negative samples.
In addition, recent researchers start to work on effective low-dimensional hyperbolic models in the KGE domain, which are detailed in Appendix~\ref{app:2}.

\subsection{A Contrastive Learning Perspective} 
Contrastive Learning has achieved remarkable success in unsupervised representation learning for image and sequential data \cite{CL-ICML2000}.
Without human supervision, Contrastive Learning methods attempt to learn the invariant representation of different views of the same instance by attracting positive pairs and separating negative pairs.
Given a training set of positive pairs $D_{\text {pos}}$, a common design of the contrastive loss is formulated as:
\begin{equation}
\mathcal{L}_{cont} = {\mathbb{E}} \left[ - {\rm log} \frac{ exp(f(x_i, x_j)/\tau)}{\sum_k exp(f(x_i, x'_k)/\tau) + exp(f(x_i, x_j)/\tau)} \right],
\label{eq:2.3}
\end{equation}
where $(x_i, x_j) \in D_{\text {pos}}$, $f(\cdot)$ is the similarity function, and $x'$ is the sampled negative instances. 
The temperature $\tau$ is a hyper-parameter, which controls the strength of penalties on different negative samples and provides a hardness-aware capability to discriminate between easy and difficult samples \cite{CL2-CVPR21}.

Recent research has proved that contrastive loss improves representation quality by optimizing two key properties asymptotically \cite{CL1-ICML20}:
(1) Alignment: to achieve the training target, two samples forming a positive pair should be assigned by similar features; and
(2) Uniformity: to preserve maximal information, feature vectors should be roughly uniformly distributed in the vector space.
We argue that, to learn invariant representations in a self-supervised way, KGE and Contrastive Learning essentially share the common properties.
Given an existing triple $(e_h,r,e_t)$, the query $q=(e_h,r)$ and the tail entity $e_t$ can be regarded as a positive pair.
The KGE model assigns the positive triple with an optimal score, which is equivalent to aligning the query vector $\mathbf{q}$ with $\mathbf{e}_t$ and separating them with the other entity vectors.

Furthermore, compared with Contrastive Learning working on images or sequential data,
KGE has its own characteristics, which inspire us to propose two innovative techniques.
First, negative samples for any entity are in a fixed range, i.e., the KG entity set, such that a more efficient strategy instead of negative sampling can be utilized to achieve a uniform vector distribution.
Second, the positive and negative samples of the entity are parameter-sharing and mutually restricted in the vector space.
Therefore, training KGE models should focus on samples that are hard to distinguish and avoid to overfit easy samples.

\section{Methodology}
\label{sec:3}

Based on the above comparative analysis between KGE and Contrastive Learning, and the newest findings in the two domains,
we propose a novel training strategy for KGE models,
namely \textbf{H}ardness-\textbf{a}ware \textbf{L}ow-dimensional \textbf{E}mbedding (\textbf{HaLE}).
Specifically, we propose a new Query Sampling loss to achieve both Alignment and Uniformity,
and to overcome the drawbacks of the traditional Negative Sampling loss, which will be detailed in Sec. \ref{sec:3.1}.
To achieve the hardness-aware ability like the temperature trick in
Contrastive Learning, we propose a novel Hardness-aware Activation mechanism in Sec. \ref{sec:3.2}.
Finally, the HaLE framework and several HaLE-based KGE models are described in Sec. \ref{sec:3.3}.

\subsection{Query Sampling Loss}
\label{sec:3.1}

Negative sampling has been proved effective and widely used to learn KG embedding and word embedding \cite{TransE, Word2vec}.
Only considering a subset of negative instances, Negative Sampling helps reduce the time complexity of one training epoch.
However, randomly sampling negative instances for each triple requires additional training time, especially for large-scale KGs.
The uncertainty in the sampling procedure brings fluctuations into KGE training and impedes model convergence \cite{RS-NSWeak}.
To omit Negative Sampling, recent work proposed two representative non-sampling approaches, i.e., All-Negative training and Non-Negative training \cite{HEKGE-WWW21,HEKGE-NAACL21}. The general loss functions of the two approaches are as follows:
\begin{align}
 \label{eq:3.1}
 \mathcal{L}_{allneg}(T) &= \underset{(e,r,e_p)\in T} {\mathbb{E}}
 \left [-log( \frac{exp(f(e,r,e_p))}{\sum\limits_{i}^{n_E}exp(f(e,r,e_i))})\right ], \\
 \label{eq:3.2}
 \mathcal{L}_{nonneg}(T) &= \underset{(e,r,e_p)\in T} {\mathbb{E}} \left[-f(e,r,e_p)\right] + g(\mathbf{E}),
\end{align}
where $T$ is the triple set of $\mathcal{G}$ and $g(\mathbf{E})$ is a regularization function.
The two approaches have respective drawbacks.
The former approach uses all entities except the target one as negative instances. It can provide a stable gradient for each epoch, while dramatically increasing computational complexity.
In the latter approach, training positive triples only can minimize computation but tends to sink the model into a trivial optimum.
Although previous work proposed some modifications to mitigate negative effects, they can only be applied to certain scoring functions and usually limit the prediction accuracy of KGE models.

Therefore, we argue that the feasible strategy replacing Negative Sampling should be somewhere between the two extreme approaches.
We first combine the two training strategies to overcome their flaws. For all training triples in $T$, we sample a small subset of triples $\hat T$ to conduct the All-Negative training, while training the rest triples via the Non-Negative approach.
Based on Eq. \ref{eq:3.1} and Eq. \ref{eq:3.2}, we can re-organize the combined loss function as:
\begin{align}
 \nonumber
 \mathcal{L} &= \mathcal{L}_{allneg}(\hat T) + \mathcal{L}_{nonneg}(T-\hat T)\\ \nonumber
 &=\underset{(e,r,e_p)\notin \hat T} {\mathbb{E}}\left[-f(e,r,e_p)\right]
 + \underset{(e,r,e_p)\in \hat T} {\mathbb{E}}\left [-log \frac{exp(f(e,r,e_p))}{\sum\limits_{i}^{n_E}exp(f(e,r,e_i))}\right ] \\ \nonumber
 &= \underset{(e,r,e_p)\in T} {\mathbb{E}}\left[-f(e,r,e_p)\right] + \underset{(e,r,e_p)\in \hat T} {\mathbb{E}}\left [log\sum\limits_{i}^{n_E}exp(f(e,r,e_i))\right ]\\
 &= -\frac{1}{n_T}\sum^{T}f(e,r,e_p) + \frac{1}{n_{\hat T}}\sum^{\hat T}LSE(f(e,r,E)),
\label{eq:3.3}
\end{align}
where the number of sampled triples $n_{\hat T} = \alpha n_T$ and $\alpha \in [0,1]$ determines the sampling proportion. $LSE(\cdot)$ is the LogSumExp function, and the regularization part in $\mathcal{L}_{nonneg}$ is omitted.

Drawing on the idea of Contrastive Learning, we can explain the intuitive sense of the reorganized loss function.
From Eq. \ref{eq:3.3}
we can see that the reorganized loss function contains two modules, which exactly satisfy two key properties of Contrastive Learning, Alignment and Uniformity.
The first module is to achieve the Alignment property by maximizing the scores of all positive triples. In the vector space,
this module maximizes the similarity of the query vector and its target vector in every pair.
The second module minimizes the similarity of each sampled query vector with all entity vectors in $\mathcal{G}$.
As all entity vectors are forced to stay away from these query vectors, the vector distribution tends to be uniform.
Focusing on the two key properties, we propose a new query sampling loss for KGE models, which is defined as:
\begin{align}
\mathcal{L} = -\frac{\lambda}{n_T}\sum^{T}f(e,r,e_p) + \frac{1}{n_{\hat T}}\sum^{\hat T}LSE(f(e,r,E)),
\label{eq:3.4}
\end{align}
where $\lambda$ is a hyper-parameter to balance the contributions of two modules. As proved by a recent work \cite{CL2-CVPR21}, learning both Alignment and Uniformity is a trade-off process.
A completely uniform vector distribution makes alignment impossible, while aligning all positive pairs perfectly causes the clustered vectors indistinguishable.
Therefore, the hyper-parameter $\lambda$ is necessary to keep KGE training stable.

A question may be raised on what is the difference between Negative Sampling and Query Sampling losses.
Negative sampling loss focuses on the relative score order of the positive triple and negative samples.
Only a few entity vectors are involved in each training batch and thus entity vectors have multifarious and ever-changing optimization directions.
In the Query Sampling loss, staying away from sampled queries are the common training target shared by all entity vectors, so most entities have a stable optimization direction.
As the goal of uniformity is clarified, we can also recognize the root flaws of the two extreme approaches.
Non-Negative training utilizes the embedding regularization to achieve a global uniformity but ignores the specific query distribution, while All-Negative training is extremely redundant to separate all query and entity vectors.
Considering above problems, our strategy samples a small proportion of queries from the query distribution to achieve a good uniformity property.

\subsection{Hardness-aware Activation Mechanism}
\label{sec:3.2}

Although our query sampling loss can keep training stable, it treats all negative instances equally.
As KGE training goes on, a large percentage of negative instances have been far away from the query vector, and thus provide less meaningful information. It would be more efficient if the loss can focus on negative instances that are difficult to be distinguished.

To solve this issue, the contrastive loss usually employs the temperature $\tau$. As shown in Eq. \ref{eq:2.3}, the feature similarities are multiplied by $\frac{1}{\tau}$ before going through the Softmax function.
As proved by the recent work \cite{CL2-CVPR21}, this temperature provides a hardness-aware ability to control the strength of penalties on hard negative samples.
However, the number of negative instances could be much higher in the All-Negative training condition. The distribution of the Softmax-normalized scores would be much more uniform, even using a higher temperature.

\begin{figure}[!tb]
\centering
\vspace{-6mm}
\includegraphics[width=0.4\textwidth]{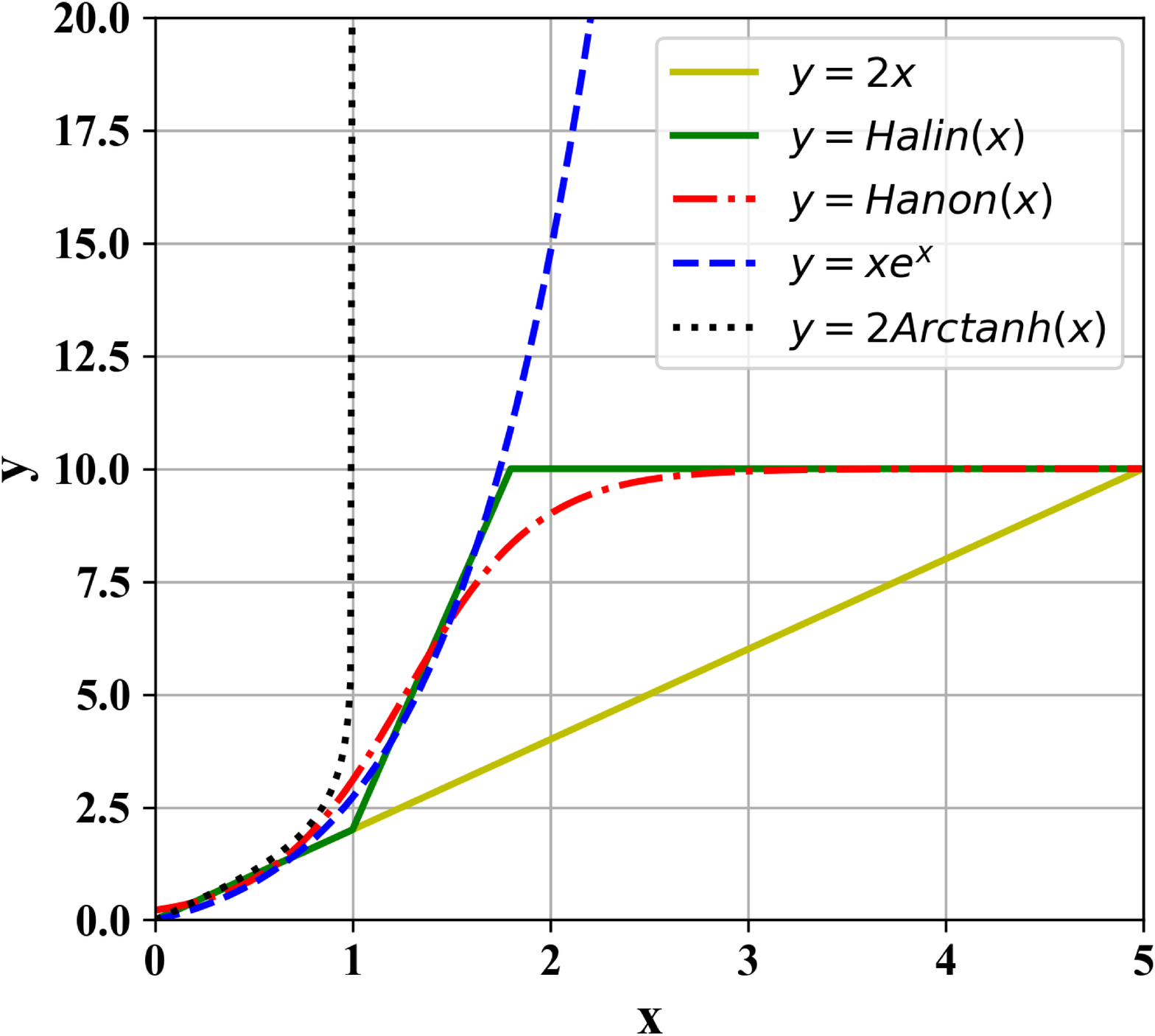}
\vspace{-4mm}
\caption{Different activation functions mentioned in this paper. The $y=2x$ is equivalent to using temperature. The $\beta$ values of $Hanon(\cdot)$ and $Halin(\cdot)$ are 3 and 10, respectively. The upper bound $\gamma$ of two functions is 10. Best viewed in color.}
\label{fig:2}
\vspace{-4mm}
\end{figure}

This problem guides us to review the recent low-dimensional hyperbolic-based KGE models \cite{MuRP,GoogleAttH}.
We find a key point overlooked in the past: the nonlinear function in the hyperbolic distance is beneficial to achieve the hardness-aware ability. As described in Appendix~\ref{app:2}, there are two nonlinear functions utilized in previous low-dimensional KGE models, i.e.,  $h(x)=2Arctanh(x)=\ln\frac{1+x}{1-x}$ and $h(x)=xe^x$.
In Fig. \ref{fig:2}, we can observe that their derivatives are always greater than one when $x \geq 0$.
Besides, they have a similar trend as the value is relatively low at first and then increases rapidly.
We argue that these nonlinear functions can be regarded as the activation function of KGE models.
Given a KGE model with $\Psi(\mathbf{q}, e)=-\|\mathbf{q}-\mathbf{e}\|_2^2$, the instance with higher $L_2$ distance is more likely to be negative.
The nonlinear activation function can further amplify the distance value of an easy negative instance, and the penalties of the indistinguishable negative instances would be strengthened in the loss.

Based on our observations, we design a Hardness-aware Activation mechanism to replace the hyperbolic geometry.
As shown in Fig. \ref{fig:2}, two existing nonlinear functions are upper unbounded, so they can only achieve a `soft constraint' to negative instances. Some easy instances that have been successfully separated are still involved in gradient computing.
There is a significant waste on large-scale KGs and it also negatively affects the alignment of the other triples.
To this end, we attempt to add a `hard constraint' by designing suitable upper-bounded activation functions.
Referring to the existing nonlinear functions, we propose two novel hardness-aware activation functions, \textbf{$Hanon(\cdot)$} and \textbf{$Halin(\cdot)$}, which are formulated as follows:
\begin{align}
&Hanon(x) = \frac{1}{\gamma^{-1}+e^{-\beta(x-0.5)}}\\
&Halin(x)=\left\{
\begin{array}{lll}
2x & & {0 < x < 1}\\
min(\beta(x-1)+2,\gamma) & & {x \geq 1 }
\end{array} \right.
\end{align}
where $\beta$ and $\gamma$ are the two hyper-parameters to control the soft and hard constraints.
We set the soft-constraint parameter $\beta$ according to the slope of $h(x)=xe^x$. The curves of the two novel functions are shown in Fig. \ref{fig:2}.

The $Hanon(\cdot)$ can be regarded as a variant of the Sigmoid function but be upper bounded by $\gamma$. The $Halin(\cdot)$ is a piecewise linear function. It has different slopes at the two sides of $x=1$ and cuts the gradients when the value is more than $\gamma$.
It is clear that $Hanon(\cdot)$ and $Halin(\cdot)$ have similar slopes to the two previous functions when $x \leq \gamma$. Then our functions use the `hard-constraint' to cut the gradients of the distinguishable instances whose distance is bigger than $\gamma$.
We further improve the Hardness-aware Activation by multiplying the similarity score with a relation-specific trainable parameter.
Applied by previous low-dimensional KGE models, this technology is beneficial to encoding hierarchical relationships and improves prediction accuracy \cite{OurRotL}.
Finally, given a query $(e,r)$ and its target entity $e_p$, the triple score based on $L_2$ distance via the Hardness-aware Activation is defined as:
\begin{align}
f_{ha}(e,r,e_p) = -h(c_r \cdot \|\Phi(e,r)-\mathbf{e}_p\|_2^2),
\end{align}
where $h(\cdot)$ denotes the activation function $Hanon(\cdot)$ or $Halin(\cdot)$, and $c_r$ is the relation-specific scalar parameter.

Note that, the Hardness-aware Activation mechanism proposed in this work
can be an alternative to the hyperbolic geometry in low-dimensional KGE models, because the former inherits the advantages of both trainable curvatures and nonlinear activation.
Besides, our solution is based on Euclidean space operations and only works in the training process, hence it is more efficient than hyperbolic KGE models.
In Sec. \ref{sec:4}, we will report our evaluation of multiple kinds of Euclidean-based KGE models using the proposed hardness-aware activation, and compare the performance of different activation functions.

\subsection{The HaLE Framework}
\label{sec:3.3}
To achieve swift and sure KGE training, the HaLE framework utilizes the Query Sampling loss to keep training stable and the Hardness-aware Activation mechanism to accelerate model convergence.
To integrate the two key techniques better, we apply the hardness-aware activation to two parts of our loss function in an asymmetric way. Specifically, we square the activated scores of positive instances in the Alignment part. Such that, the positive instances would get much stricter regularization than negative ones. The false-negative instances would get bigger gradients than negative instances, and a positive instance whose $L_2$ distance is close to zero would make less contribution to the loss. We find that it can accelerate the model convergence further. Therefore, the final loss function of the HaLE framework is formulated as follows:
\begin{align}
 \mathcal{L}_{HaLE} &= -\frac{\lambda}{n_T}\sum^{T}f_{ha}(e,r,e_p)^2 + \frac{1}{n_{\hat T}}\sum^{\hat T}LSE(f_{ha}(e,r,E)),
\label{eq:3.4}
\end{align}

To verify the performance of our HaLE framework, we select five representative KGE models: TransE \cite{TransE}, DistMult \cite{DistMult}, RotatE \cite{RotatE}, RotE \cite{GoogleAttH}, and RotL \cite{OurRotL}.
These models utilize five different transform functions to generate the query vector in the Euclidean space, which are formulated as follows:
\begin{align}
 TransE: &\Phi_T(e_h,r)=\mathbf{e}_h+\mathbf{r}, \\
 DistMult: &\Phi_D(e_h,r)=\mathbf{e}_h \cdot \mathbf{r}, \\
 RotatE: &\Phi_R(e_h,r)=Rot(\mathbf{r},\mathbf{e}_h), \\
 RotE: &\Phi_E(e_h,r)=Rot(\mathbf{r},\mathbf{e}_h) + \mathbf{r'}, \\
 RotL: &\Phi_L(e_h,r)=Rot(\mathbf{r},\mathbf{e}_h) \oplus_\alpha \mathbf{r'},
\end{align}
where $Rot(\cdot)$ denotes the vector rotation operation, $\mathbf{r}$ and $\mathbf{r'}$ are two different relation vectors corresponding to the same relation $r$.
To make a fair comparison, we
use the $L_2$-distance squared similarity function $\Psi(\mathbf{q}, e)=-\|\mathbf{q}-\mathbf{e}\|_2^2$. Although some previous works use the dot product function with specific regularization items, it has been proven to be equivalent to $L_2$ distance squared \cite{DualRegularizer}.

Compared with the previous Negative Sampling, All-Negative, and Non-Negative approaches, our HaLE framework can achieve swift and sure KGE training for several reasons:
\begin{itemize}
 \item The training process in one epoch is greatly accelerated. The negative sampling for each triple is omitted, and the query sampling can significantly reduce the All-Negative training cost.
 \item HaLE can provide a stable training target.
 In each step of parameter optimization,
 we compute the gradients of all positive triples and force all entity vectors to stay away from the same part of queries.
 \item The total training time is reduced. The new loss can avoid parameter fluctuation, and the hardness-aware activation can focus on difficult instances.
 As a result, the HaLE-trained model can converge quickly in several epochs.
\end{itemize}

\section{Experiments}
\label{sec:4}

\subsection{Experimental Setup}
To verify the performance of HaLE, we focus on the link prediction, the most typical and challenging task for KG embeddings.
Different from previous KGE research efforts that pursuing a higher prediction accuracy, we concentrate
on the {\em training efficiency} of KGE models, which is critical for them to be applied in practice.

To compare the training efficiency of different strategies, we employ five representative KGE models as mentioned in Sec. \ref{sec:3.3} and train them in the specific space and time conditions.
For the space condition, we set the dimension number of the low-dimensional models as 32 and high-dimensional ones as 256.
For the time condition, we set a maximum training time according to the KG size of each dataset, as shown in Table \ref{tab:2}.
Following the previous work \cite{TransE}, we adopt two kinds of evaluation metrics in the `Filter' mode:
(1) MRR, the average inverse rank of the test triples, and
(2) Hits@N, the proportion of correct entities ranked in top N.
Higher MRR and Hits@N mean better performance.

\begin{table}[h]
\vspace{-2mm}
\caption{Statistics of the datasets.}
\centering
\small
\label{tab:2}
\vspace{-2mm}
\begin{tabular}{c|rrrrrr}
\hline
\textbf{Dataset} & \textbf{$n_R$} & \textbf{$n_E$} & \textbf{\#Train} & \textbf{\#Valid} & \textbf{\#Test} & \textbf{Time} \\
\hline
FB15k237 & 237 & 14,541 & 272,115 & 17,535 & 20,466 & 1,200s\\
WN18RR & 11 & 40,943 & 86,845 & 3,034 & 3,134 & 600s\\
CoDEx-S	& 42 & 2,034 & 32,888 & 1,827 & 1,828 & 300s\\
CoDEx-M	& 51 & 17,050 &	185,584 & 10,310 & 10,311 & 1,200s\\
CoDEx-L	& 69 & 77,951 &	551,193 & 30,622 & 30,622 & 1,200s \\
\hline
\end{tabular}
\vspace{-3mm}
\end{table}

\subsubsection{Datasets}
Our experimental studies are conducted on five commonly used datasets.
WN18RR \cite{WN18RR} is a subset of the English lexical database WordNet.
FB15k237 \cite{FB15k237} is extracted from Freebase including knowledge facts about movies, actors, awards, and sports. Compared with the FB15k dataset, it removes inverse relations because many test triples can be obtained simply by inverting triples in the training set.
CoDEx-S/M/L \cite{Codex} are three KG datasets with different scales extracted from Wikidata and Wikipedia. We only use positive triples in each dataset for a fair comparison.
The statistics of the datasets are given in
Table \ref{tab:2}.
`Train', `Valid', and `Test' refer to the number of triples in the training, validation and test sets.

\begin{table*}[!tb]
\caption{Low-dimensional link prediction results on the WN18RR and FB15k237 datasets. The symbol `$^*$' means the model is fully-trained, otherwise the model is trained in limited time. `Cost' means training time (minutes). The best score of fully-trained models \underline{underlined} and the best score of limited-trained models in \textbf{Bold}.}
\centering
\small
\label{tab:3}
\begin{tabular}{cl|cccc|cccc}
\toprule
\multirow{2}*{\textbf{Type}} & \multirow{2}*{\textbf{Methods}} & \multicolumn{4}{c|}{\textbf{FB15K237}} & \multicolumn{4}{c}{\textbf{WN18RR}}\\
~ & ~ & \textbf{MRR} & \textbf{Hits@10} & \textbf{Hits@1}  & \textbf{Cost} & \textbf{MRR} & \textbf{Hits@10} & \textbf{Hits@1} & \textbf{Cost} \\
\midrule
\multirow{4}*{\makecell[c]{Fully-trained \\ Hyperbolic-based \\ Models}} & MuRP \cite{MuRP}* & .323 & \underline{.501} & .235 & $335$ & .465 & .544 & .420 & $117$\\
~ & RefH \cite{GoogleAttH}* & .312 & .489 & .224 & $252$ & .447 & .518 & .408 & $84$ \\
~ & RotH \cite{GoogleAttH}* & .314 & .497 & .223 & $242$ & \underline{.472} & \underline{.553} & \underline{.428} & $82$ \\
~ & AttH \cite{GoogleAttH}* & \underline{.324} & \underline{.501} & \underline{.236} & $276$ & .466 & .551 & .419 & $94$ \\
\hline
\multirow{5}*{\makecell[c]{Limited Time \\Negative Sampling \\ Trained}} & TransE \cite{TransE} & .243 & .422 & .154  & $20$ & .177 & .417 & .045   & $10$\\
~ & DistMult \cite{DistMult} & .278 & .445 & .194  & $20$ & .351 & .482 & .283 & $10$ \\
~ & RotatE \cite{RotatE} & .223 & .391 & .141  & $20$ & .346 & .460 & .285 & $10$\\
~ & RotE \cite{GoogleAttH} & .246 & .424 & .159  & $20$ & .355 & .480 & .290 & $10$\\
~ & RotL \cite{OurRotL} & .140 & .266 & .079 & $20$ & .295 & .368 & .254 & $10$\\
\hline
\multirow{5}*{\makecell[c]{Limited Time \\HaLE Trained}} & TransE \cite{TransE} & .314 & .492 & .224 & $20$ & .212 & .492 & .028  & $10$ \\
~ & DistMult \cite{DistMult} & .308 & .483 &  .222  & $20$ & .447 & .533 & .399  & $10$ \\
~ & RotatE \cite{RotatE} & .307 & .479 & .219  & $20$ & .451 & .536 & .406  & $10$ \\
~ & RotE \cite{GoogleAttH} & .313 & .486 & .226  & $20$ & .460 & .542 & .416  & $10$ \\
~ & RotL \cite{OurRotL} & \textbf{.316} & \textbf{.493} & \textbf{.228} & $20$ & \textbf{.471} & \textbf{.558} & \textbf{.424} & $10$\\
\bottomrule
\end{tabular}
\end{table*}

\subsubsection{Comparing Methods}
\label{sec:4.1.2}
We compare different training strategies mentioned in
Sec. \ref{sec:3}, including Negative Sampling
(SamNeg)
\cite{GoogleAttH}, Self-Adversarial Negative Sampling
(AdvNeg) \cite{RotatE}, All-negative Training
(AllNeg) \cite{HEKGE-WWW21} and Non-negative Training
(NonNeg) \cite{HEKGE-NAACL21}.
SamNeg and AdvNeg
utilize the binary cross entropy loss, while AllNeg utilizes the cross-entropy loss to compute all candidate entities. We implement a general NonNeg strategy which uses a square loss to maximize positive triple scores and a global regularization to constrain the distance between each entity vector and the center vector of entity matrix.
In the HaLE framework, we use the activation function $Hanon(\cdot)$ by default. We also compare multiple activation functions shown in the Fig. \ref{fig:2}.

\subsubsection{Implementation Details}
We select the hyper-parameters of our model via grid search according to the metrics on the validation set.
For previous strategies, we select the learning rate among $\{0.0005, 0.001, 0.005\}$, the number of negative samples among $\{50, 256, 512\}$, the batch size among $\{256, 512, 1,024\}$.
For the HaLE framework, we select the sampling proportion $\alpha$ among $\{0.05, 0.1, 0.2\}$, the balancing ratio $\lambda$ among $\{0.1,0.3,0.5,1.0\}$, the hard-constraint parameter $\gamma$ among $\{5, 10, 20\}$.
All experiments are performed on Intel Core i7-7700K CPU @ 4.20GHz and NVIDIA GeForce GTX1070 GPU, and are implemented in Python using the PyTorch framework.

\subsection{Experimental Results}

As Negative Sampling is the most commonly used training strategy in the KGE domain, we first compare the limited-time performance of different KGE models trained by Negative Sampling and HaLE (with Hanon).
Meanwhile, we provide public results of several fully-trained KGE models and measure their training time via the corresponding open-source codes.
The 32-dimensional results on WN18RR and FB15k237 are shown in Table \ref{tab:3}, while 256-dimensional results on three CoDEx datasets are shown in Table \ref{tab:4}.
Due to space constraint, the other experimental results including the ablation experiments and the visualization of entity embeddings can be found in Appendix~\ref{app:3} and \ref{app:4}.

\subsubsection{Low-dimensional Performance Comparison}
From Table \ref{tab:3}, we have the following observations.
Setting the limited training time as 20 minutes for FB15k237 and ten minutes for WN18RR, the five different models trained by HaLE significantly outperform the ones trained by Negative Sampling (SamNeg)
on both datasets. The MRR and Hits@10 of all models have an average 5\% increase. The results indicate the effectiveness of the HaLE framework.
In the five models, the HaLE-trained RotL model achieves the best performance in all metrics on two datasets, HaLE-RotE is the second. It proves the effectiveness of the rotation-translation form in the transform function.
In addition, we find that the SamNeg-trained RotL model is weaker than others, because the effect of the flexible addition operation relies on the nonlinear activation, which does not exist in the normal $L_2$-distance similarity function.
Compared with fully-trained hyperbolic-based models, the simplest TransE model achieves competitive performance on FB15k237 after being trained in less than 20 minutes by HaLE. The HaLE-trained RotL model obtains the state-of-the-art MRR and Hits@10 on WN18RR, which has no hyperbolic geometry and only costs less than 10 minutes.
The results indicate that HaLE can improve the Euclidean-base models in low-dimensional conditions, only spending around one tenth of the training time of the hyperbolic-based models.

\subsubsection{High-dimensional Performance Comparison}
In the high-dimensional condition, HaLE shows a significant advantage over Negative Sampling.
In the limited training time, HaLE can accelerate model convergence while
SamNeg-trained models fail due to the unstable gradients. This difference is more significant in the large-scale KGs.
On the CoDEx-L dataset, the performance of TransE trained by HaLE is almost SamNeg-trained TransE.
Table \ref{tab:4} also lists the results of five fully-trained KGE models using more than 256 dimensions, which are detailed-tuned by a powerful hyperparameter optimization method \cite{Codex}.
From the table, we can see that the HaLE-trained models show strong competitiveness. Especially on CoDEx-M, the limited-time
trained HaLE-TransE model (trained in 20 minutes) outperforms that of the 512-dimensional fully-trained TransE model (trained for more than one hour, 77 minutes).
Although the Hits@1 of the optimal HaLE-RotL model is still lower than those benchmarks on CoDEx-L, HaLE-RotL achieves great Hits@10 results using less training time and fewer parameters.
Training in a limited time, HaLE-trained models already obtain similar performance to the state-of-the-art models on five datasets.
These results prove the efficiency of our HaLE framework on keeping high prediction accuracy.

\begin{table*}[!tb]
\caption{High-dimensional link prediction results on the CoDEx datasets. The symbol `$^*$' means the model is fully-trained, otherwise the model is trained in limited time. `SamNeg-' means the model is trained by Negative Sampling, and `Cost' means training time (minutes).
The best score of fully-trained models \underline{underlined} and the best score of limited-trained models in \textbf{Bold}.}
\centering
\small
\label{tab:4}
\begin{tabular}{l|cccc|cccc|cccc}
\toprule
\multirow{2}*{\textbf{Methods}} & \multicolumn{4}{c|}{\textbf{CoDEx-S}} & \multicolumn{4}{c|}{\textbf{CoDEx-M}} & \multicolumn{4}{c}{\textbf{CoDEx-L}}\\
~ & \textbf{MRR} & \textbf{Hits@10} & \textbf{Hits@1} & \textbf{Cost} & \textbf{MRR} & \textbf{Hits@10} & \textbf{Hits@1} & \textbf{Cost} & \textbf{MRR} & \textbf{Hits@10} & \textbf{Hits@1} & \textbf{Cost}\\
\midrule
RESCAL \cite{Rescal}* & .404 & .623 & .293 & $10$ & .317 & .456 & .244 & $34$ & .304 & .419 & .242 & $67$ \\
TransE \cite{TransE}* & .354 & .634 & .219 & $9$  & .303 & .454 & .223 & $77$ & .187 & .317 & .116 & $34$ \\
ComplEx \cite{ComplEx}* & \underline{.465} & \underline{.646} & \underline{.372} & $6$ & \underline{.337} & \underline{.476} & \underline{.262} & $87$ & .294 & .400 & .237 & $50$ \\
ConvE \cite{ConvE}*  & .444 & .635 & .343 & $9$ & .318 & .464 & .239 & $139$  & .303 & .420 & .240 & $688$ \\
TuckER \cite{TuckER}* & .444 & .638 & .339 & $39$ & .328 & .458 & .259 & $152$ & \underline{.309} & \underline{.430} & \underline{.244} & $440$ \\
\hline
SamNeg-TransE \cite{TransE} & .301 & .544 & .177 & $5$ & .178 & .327 & .107 & $20$ & .144 &  .260 & .086 & $20$ \\
SamNeg-DistMult \cite{DistMult} & .360 & .589 & .246 & $5$ & .255 & .395 & .182 & $20$ & .228 &  .353 & .164 & $20$ \\
SamNeg-RotatE \cite{RotatE} & .327 & .546 & .214 & $5$ & .182 & .327 & .110 & $20$ & .159 &  .281 & .099 & $20$ \\
SamNeg-RotE \cite{GoogleAttH} & .328 & .549 & .214 & $5$ & .183 & .330 & .112 & $20$ & .155 &  .270 & .097 & $20$\\
SamNeg-RotL \cite{OurRotL}  & .313 & .534 & .205 & $5$ & .162 & .259 & .106 & $20$ & .055 &  .113 & .026 & $20$ \\
\hline
HaLE-TransE \cite{TransE} & .353 & .620 & .223 & $5$ & .313 & .467 & .230 & $20$ & .300 &  .436 & .226 & $20$ \\
HaLE-DistMult \cite{DistMult} & .403 & .629 & .289 & $5$ & .314 & .462 & .236 & $20$ & .299 &  .427 & .230 & $20$ \\
HaLE-RotatE \cite{RotatE} & .407 & .635 & .289 & $5$ & .324 & .474 & .244 & $20$ & .302 &  .435 & .229 & $20$ \\
HaLE-RotE \cite{GoogleAttH} & \textbf{.409} & \textbf{.639} & .291 & $5$ & \textbf{.326} & \textbf{.475} & \textbf{.246} & $20$ & \textbf{.308} &  \textbf{.438} & .237 & $20$ \\
HaLE-RotL \cite{OurRotL} & .408 & \textbf{.639} & \textbf{.292} & $5$ & .324 & .474 & .244  & $20$& \textbf{.308} &  \textbf{.438} & \textbf{.238} & $20$ \\
\bottomrule
\end{tabular}
\vspace{-2mm}
\end{table*}

\subsubsection{Efficiency Comparison for Query Sampling}
To verify the training efficiency of HaLE, we select the 32-dimensional RotE model and compare Query Sampling loss with four previous training strategies
as mentioned in Sec. \ref{sec:4.1.2}.
The performance changes of the validation set as training proceeds are shown in the three upper line charts in Fig. \ref{fig:3}.
It is clear that our HaLE achieves remarkable efficiency on the three datasets comparing with the previous strategies.
Besides, there are some common observations in the three results.
Except NonNeg, SamNeg (Negative Sampling) is the worst one whose Hits@10 slowly increases in the first 50 seconds, indicating the negative effect of unstable gradients.
Assigning different weights to negative instances, AdvNeg has a much faster convergence speed than SamNeg. Outperforming SamNeg and
AdvNeg, the AllNeg strategy has good performance on two large datasets FB15k237 and CoDEx-M.
As the WN18RR is relatively sparse, AllNeg is slightly inefficient to train all negative instances using a uniform gradient.
These results prove the effectiveness of the All-Negative training and hardness-aware ability.
Without negative instances, the NonNeg strategy is the only unstable one. Keeping increasing in the first 80 seconds, the Hits@10 of NonNeg starts decreasing, because its negative constraint is not powerful enough to avoid over-fitting.
Our HaLE achieves the fastest convergence speed on the three datasets. Especially on WN18RR, HaLE achieves more than 0.5 Hits@10 in the first 30 seconds, which is already better than the final results of the others, indicating that HaLE can achieve a swift and sure KGE training, and has considerable potential in practical applications.

\subsubsection{Efficiency Comparison for Hardness-aware Activation}
\begin{figure*}[!bt]
\centering
\includegraphics[width=0.9\textwidth]{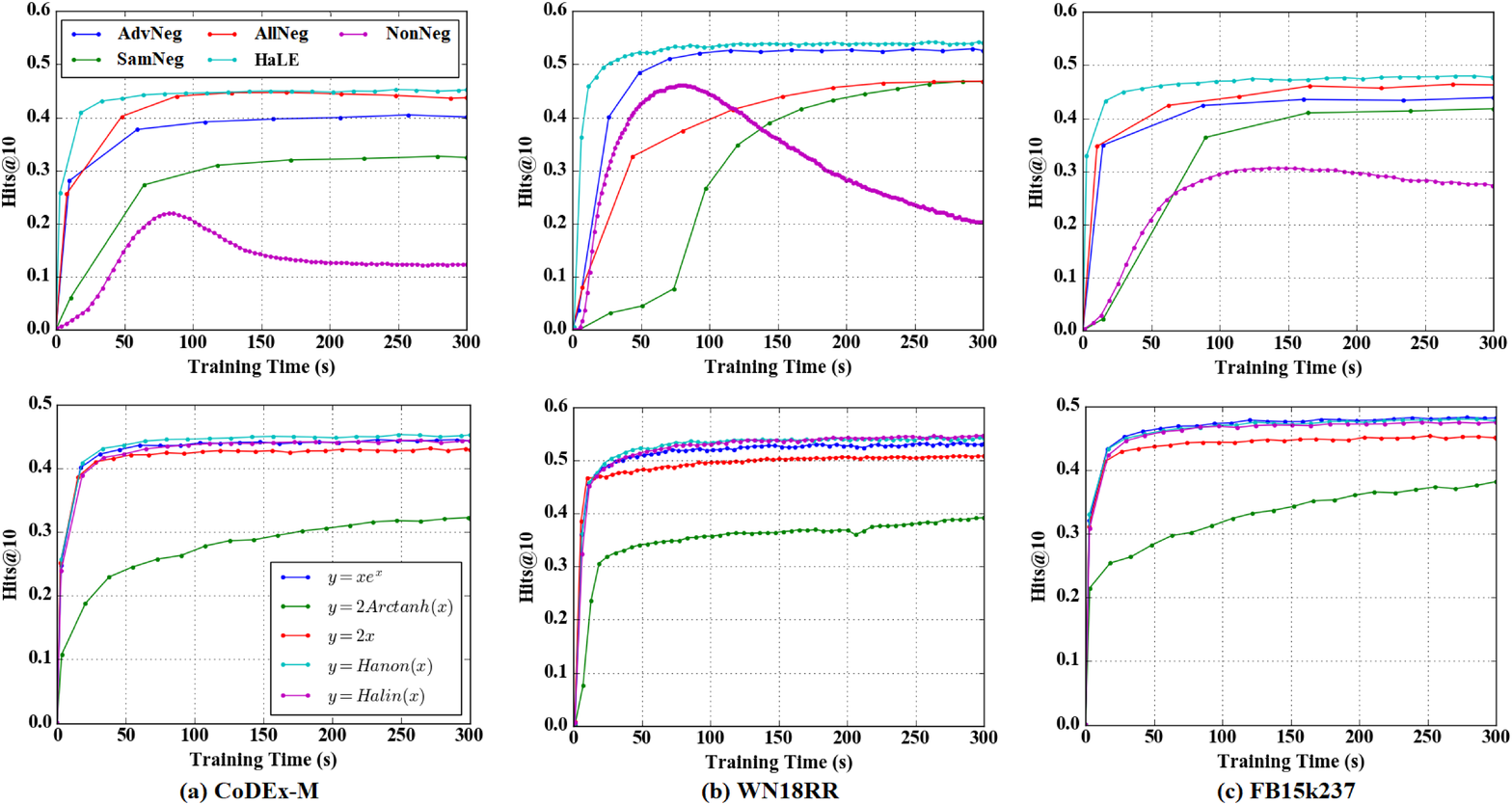}
\vspace{-4mm}
\caption{The Hits@10 of 32-dimensional RotE-based model as training proceeds on three datasets. The upper charts compare different loss functions, while the lower charts compare different activation functions. Best viewed in color.}
\label{fig:3}
\vspace{-2mm}
\end{figure*}

To verify the hardness-aware activation mechanism and our novel activation functions, we compare the performance of different activation functions in the HaLE framework. The results are shown in the three lower line charts in Fig. \ref{fig:3}.
We can find similar observations on the three datasets.
At first, utilized in hyperbolic models, the Arctanh-based function is obviously weaker than others. As its definition range is $(-1,1)$, it relies on the normalization effect of hyperbolic geometry.
In the two linear functions, the $y=2x$ function is simulating the temperature control in Contrastive Learning and our $Halin(\cdot)$ can be regarded as an extended version of the former with soft- and hard-constraints.
It is clear that the Hits@10 of $y=2x$ increases faster in the first few rounds but gets a lower final Hits@10.
Then we can see that the performance of $Hanon(\cdot)$, $Halin(\cdot)$ and $y=xe^x$ are very close.
On the CoDEx-M and WN18RR datasets, our Hanon function performs the best. It is because $Hanon(\cdot)$ has an additional hard-constraint property to limit the outputted maximum. The $y=xe^x$ function used in the original RotL model is slightly better than $Hanon(\cdot)$ in FB15k237.
The linear function $Halin(\cdot)$ achieves similar performance with the two nonlinear ones,
proving that the soft- and hard-constraints are the key properties of the hardness-aware activation.

\section{Related Work}
\label{sec:5}

Recently, knowledge graph completion via KGE has been an active research topic \cite{2017Survey}. Dozens of KGE models have been proposed, which can be divided into three categories from the perspective of the scoring function: (1) geometric distance based models, including TransE \cite{TransE}, TransD \cite{TransD}, RotatE \cite{RotatE}, QuatE \cite{QuatE}; (2) tensor factorization based models, including RESCAL \cite{Rescal}, DistMult \cite{DistMult}, ComplEx \cite{ComplEx}, TuckER \cite{TuckER}; and (3) deep learning based models, including ConvE \cite{ConvE}, ConvKB \cite{ConvKB}, RGCN \cite{RGCN}, SACN \cite{SACN}, CompGCN \cite{CompGCN}.
All current KGE models suffer from the same issues of low speed and high cost in the training phase.
The problem become much more serious when processing large-scale KGs with millions or billions of entities.
Recently, several researchers have worked on this issue via different technical channels.

\vspace{2mm}
\noindent\textit{Reducing Parameters.}
Limiting the vector dimensions as 32 or 64, several low-dimensional KGE models are proposed to achieve competitive performance
with less trainable parameters.
MuRP \cite{MuRP} is the first KGE model based on hyperbolic vector space, and outperforms previous models in the low-dimensional condition. It embeds KG triples in the Poincar\'{e} ball model using the M\"{o}bius matrix-vector multiplication and M{\"o}bius addition operations.
To further capture logical patterns in KGs, Chami et al. \cite{GoogleAttH} propose a series of hyperbolic KGE models, including RotH, RefH, and AttH. These models utilize vector rotation or reflection operations to replace the multiplication operation between the head entity and relation vectors.
Based on the RotH model, Wang et al. \cite{OurRotL} propose two Euclidean-based lightweight models, RotL and Rot2L. Eliminating complex hyperbolic operations, the two models have lower computational complexity and faster convergence speed.

\vspace{2mm}
\noindent\textit{Replacing Negative Sampling.}
Most current KGE models are trained via Negative Sampling, which considers only a subset of negative instances to reduce the time complexity of each training epoch.
However, Negative Sampling usually extends the convergence time of KGE models because of additional sampling calculations and unsteady training gradients.
To solve this issue, Li et al. \cite{HEKGE-WWW21} propose an efficient All-Negative training framework and reduce the complexity of All-Negative calculations by dis-entangling the interactions between entities. However, this framework can only be applied to KGE models using a square-based loss. Its accuracy is lower than models trained by the Negative Sampling loss, especially in the low-dimensional condition.
Peng et al. \cite{HEKGE-NAACL21} employ segmented embeddings for parallel processing, and propose a Non-Negative strategy utilizing two vector constraints to replace Negative Sampling.
However, these techniques cause a decrease in accuracy and force the model to use higher-dimensional vectors to represent each entity (e.g., 2,000 dimensions in \cite{HEKGE-NAACL21}). Besides, this framework is based on Orthogonal Procrustes Analysis, and cannot be applied to existing KGE models directly.

\vspace{2mm}
\noindent\textit{Accelerating Model Convergence.}
Some recent research efforts design new loss functions to accelerate the convergence of KGE models.
Sun et al. \cite{RotatE} propose a self-adversarial negative sampling technique for efficiently and effectively training the RotatE model. It can be regarded as an improved binary cross-entropy loss, treating the normalized triple score as the weight of each negative sample.
Another way is adding soft label loss based on the knowledge distillation technique.
DistilE \cite{DistilE} utilizes an additional soft-label loss based on the knowledge distillation technique. A pre-trained high-dimensional KGE model generates soft labels for each training sample and accelerates the convergence of the small student model.
MulDE \cite{OurMulDE} is a multi-teacher knowledge distillation framework for KGE models. Instead of a high-dimensional model, MulDE employs multiple low-dimensional models as teachers jointly supervising the student model via a novel iterative distillation strategy.
Although the knowledge distillation framework can train a student KGE model quickly, it still requires the pre-training of teacher models and cannot really reduce training cost.


In this paper, we focus on the lightweight training of KGE models, which holds great potential for realizing a lifelong iterative process of
many important Web applications such as Web search and recommendations.
KGE can provide semantically-rich representations for entities, which can enhance the information extraction models to extract new knowledge triples from the Web.
As HaLE effectively reduces the training time,
KGE models can be rapidly updated to support iterative processing.

\vspace{-2mm}
\section{Conclusion}
\label{sec:6}
Recent knowledge graph embedding (KGE) models excessively pursue prediction accuracy but
ignore the training efficiency.
In this paper, we propose a novel Hardness-aware Low-dimensional Embedding (HaLE) framework to achieve a swift and sure KGE training. Motivated by the newest findings in the Contrastive Learning domain, we propose two key techniques: {\em Query Sampling Loss} and {\em Hardness-aware Activation}. We describe the connections of the two techniques with previous KGE achievements and prove their effectiveness by
comparing with four previous training strategies in the link prediction task.
The experimental results show that HaLE can achieve both higher prediction accuracy and faster convergence speed in limited training time.

These positive results encourage us to explore further research activities in the future.
Instead of using artificially designed activation functions, we will apply the Neural Architecture Search technology to find more powerful activation functions automatically.
Facing large-scale KGs, All-Negative training is still a burden. We aim to filter out some negative instances before measuring scores based on the hard-constraint mechanism.
Finally, we will apply HaLE to other time-consuming KGE models such as ConvE \cite{ConvE} and SACN \cite{SACN}, and more KG tasks such as KG alignment and triple classification, to further verify the performance of the proposed framework.

\vspace{-2mm}
\begin{acks}
This research is partially supported by the National Natural Science Foundation in China
(Grant: 61672128) and the Fundamental Research Fund for Central University (Grant: DUT20TD107).
Quan Z. Sheng is partially supported by Australian Research Council (ARC) Future Fellowship Grant FT140101247, and Discovery Project Grant DP200102298.
Kai Wang would like to thank the incomparable love and support from Dan Lin over the past decade. 
No matter how hard life may get unawares, his love for her is sweet and sure.
He also would like to thank Macquarie University for offering the Cotutelle PhD Scholarship.
\end{acks}

\bibliographystyle{ACM-Reference-Format}
\bibliography{bibtex}
\clearpage

\appendix
\section{Appendix}

\subsection{Summary of Notations}
\label{app:1}
The main notations used in this paper and their descriptions are summarized in Table~\ref{tab:1}.

\begin{table}[h]
\caption{Summary of the notations in this paper.}
\centering
\label{tab:1}
\begin{tabular}{c|l}
\hline
\textbf{Symbol} & \textbf{Description} \\[3pt]
\hline
$\mathcal{G}$ & A knowledge graph (KG) \\[3pt]
$T$ & The set of existing triples in a KG\\[3pt]
$E,R$ & The set of entities (E) or relations (R) in a KG\\[3pt]
$n_T, n_E, n_R$ & The item number in a specific set\\ [3pt]
$e,r$ & An entity (e) or a relation (r) in a KG\\[3pt]
$q=(e,r)$ & A query containing an entity and a relation \\[3pt]
$e_p$ & The target entity of a query \\[3pt]
$\mathbf{e}, \mathbf{r}$ & Embedding vector of the entity e or the relation r\\[3pt]
$\mathbf{q}$ & Embedding vector of the query q\\[3pt]
$d$ & Dimension of the embedding vectors\\[3pt]
$f(e_h,r,e_t)$ & The scoring function of a KGE model\\[3pt]
$\Phi(e_h,r)$ & The transform function of a KGE model\\[3pt]
$\Psi(q,e_t)$ & The similarity function of a KGE model\\[3pt]
$h(s)$ & The activation function for triple scores\\[3pt]
\hline
\end{tabular}
\end{table}

\subsection{Hyperbolic KGE Models}
\label{app:2}

Recently, researchers have proposed effective low-dimensional KGE models based on hyperbolic geometry, such as MuRP, RotH, RefH and AttH \cite{MuRP,GoogleAttH}.
These hyperbolic KGE models initialize the entity embedding vectors in the $d$-dimensional Poincar\'{e} ball \cite{HBMath2} with negative curvature $\textnormal{-}c$ $(c \textgreater 0)$: $\mathbb{B}_c^d=\{\mathbf{x} \in \mathbb{R}^{d}:\|\mathbf{x}\|^{2}<\frac{1}{c}\}$, where $\| \cdot \|$ denotes the $L_2$ norm.
The transform functions employed are similar to previous Euclidean-based models, but apply the M{\"o}bius addition and M{\"o}bius matrix-vector multiplication in the hyperbolic space.
The similarity function is the key component of hyperbolic models.
They utilize the hyperbolic distance to measure the similarity among entity vectors, which is defined as:
\begin{align}
&\Psi_{hyp}(\mathbf{q}, e_p) = -\frac{2}{\sqrt{c}}Arctanh(\sqrt{c}\|-\mathbf{q} \oplus_c \mathbf{e}_p\|)^2, \label{eq:A.1}
\end{align}
where $Arctanh(x) = \frac{1}{2}\ln\frac{1+x}{1-x}$, and $\oplus_c$ is the M{\"o}bius addition operation.
To eliminate the complicated calculations in hyperbolic space, Wang et al. \cite{OurRotL} analyzed the effect of hyperbolic geometry and proposed a lightweight Euclidean-based model RotL, whose similarity function is defined as:
\begin{align}
&\Psi_{euc}(\mathbf{q}, e_p) = -\varphi \left(\left\|\frac{\alpha(-\mathbf{q} + \mathbf{e}_p)}{1-\mathbf{q} \cdot \mathbf{e}_p}\right\|\right)
\label{eq:A.2}
\end{align}
where $\alpha$ is a relation-specific trainable parameter, and the Arctanh function is replaced by a simpler nonlinear function $\varphi(x)=xe^x$.

\subsection{Ablation Experimental Results}
\label{app:3}

We make a series of ablation experiments to evaluate different modules in HaLE. Specifically, we compare the performance of two training strategies after using our hardness-aware activation mechanism.
In Table \ref{tab:5}, `Activation' and `RelRatio' refer to the activation function and the trainable relation-specific parameters, while `PosSquare' refers to using the square of positive scores.
`HA + SamNeg' and `HA + AllNeg' represent applying
hardness-aware activation to the SamNeg and AllNeg training strategies, respectively.
The results prove the effectiveness of the three major modules in hardness-aware activation, and indicate that this mechanism can also increase the performance of Negative Sampling and All-Negative training strategies.

\begin{table}[h]
\caption{The results of ablation experiments on the 32d HaLE-RotL model.}
\centering
\label{tab:5}
\begin{tabular}{lllrrr}
\toprule
\textbf{Dataset} & \textbf{Methods} &  \textbf{MRR} & \textbf{Hits@10} & \textbf{Hits@1} \\
\midrule
\multirow{6}*{CodexM} &  HaLE-RotL & \textbf{0.309} & \textbf{0.454} &   \textbf{0.231} \\
~ & w/o Activation  & 0.289 &    0.436 &   0.212 \\
~ &    w/o PosSquare  & 0.241 &    0.351 &   0.185 \\
~ & w/o RelRatio  & 0.293 &    0.438 &   0.214 \\
~ &    HA + SamNeg  & 0.297 &    0.443 &   0.219 \\
~ &    HA + AllNeg  & 0.262 &    0.415 &   0.182 \\
\hline
\multirow{6}*{FB15k237} &  HaLE-RotL & \textbf{0.316} & \textbf{0.493} & \textbf{0.228} \\
~ & w/o Activation  & 0.294 &    0.456 &   0.212 \\
~ &    w/o PosSquare  & 0.268 &    0.405 &   0.202 \\
~ & w/o RelRatio  & 0.295 &    0.460 &   0.212 \\
~ &    HA + SamNeg  & 0.314 &    0.490 &   0.225 \\
~ & HA + AllNeg  & 0.287 &    0.454 &   0.204 \\
\hline
\multirow{6}*{WN18RR} &  HaLE-RotL & \textbf{0.471} &  \textbf{0.558} &  \textbf{0.424} \\
~ & w/o Activation  & 0.426 &    0.520 &   0.376 \\
~ &    w/o PosSquare  & 0.469 &    0.558 &   0.420 \\
~ & w/o RelRatio  & 0.454 &    0.548 &   0.404 \\
~ &    HA + SamNeg  & 0.457 &    0.553 &   0.394 \\
~ & HA + AllNeg & 0.449 &    0.540 &   0.387 \\
\bottomrule
\end{tabular}
\end{table}

\subsection{Visualization of Entity Embeddings Distribution}
\label{app:4}
\begin{figure*} 
\centering
\includegraphics[width=1\textwidth]{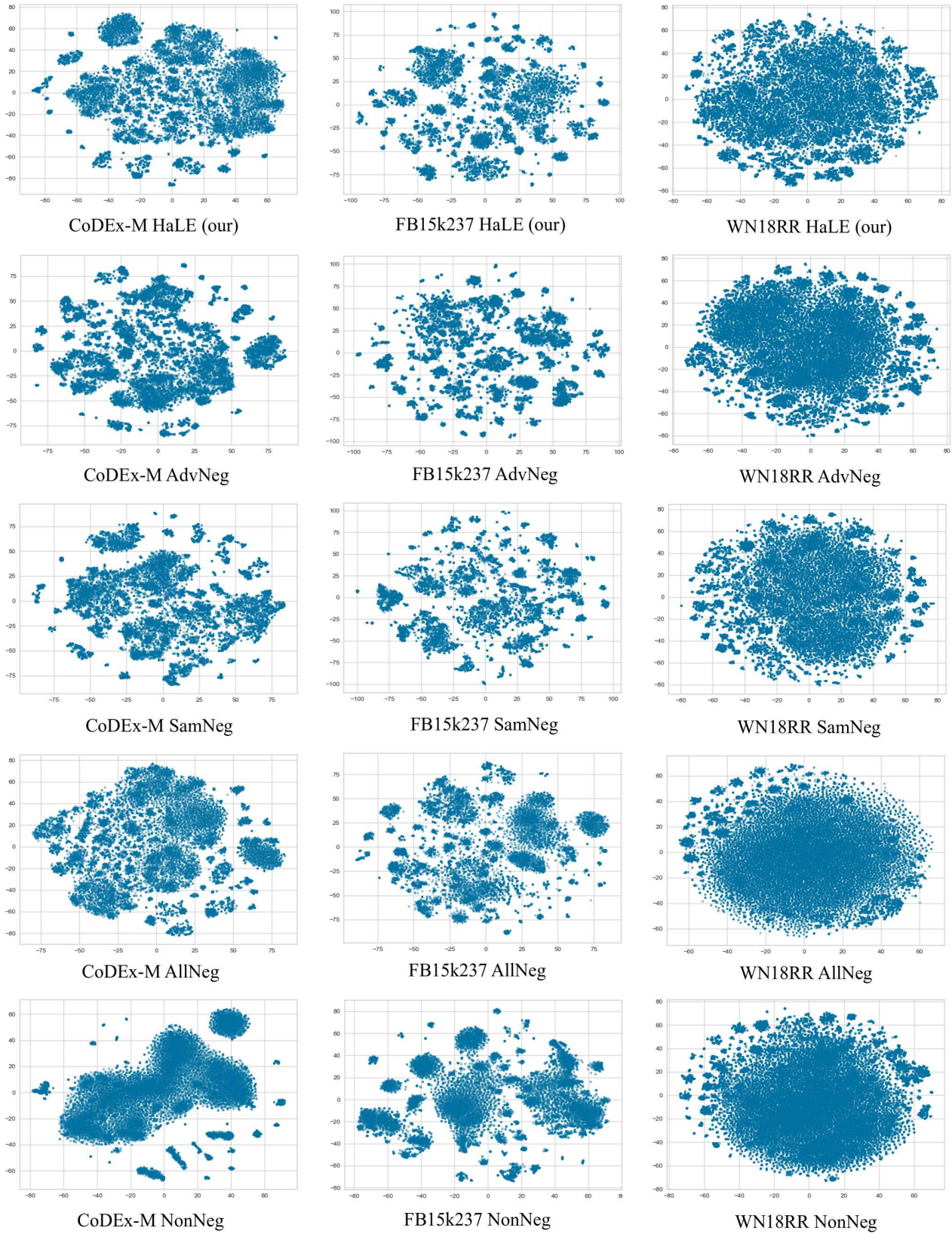}
\caption{The visualization of entity embeddings of KGE models trained by different strategies. The vector distributions are generated by t-SNE.}
\label{fig:4}
\end{figure*}

We also visualize the vector distribution of entity embeddings using the
t-SNE dimensionality reduction method.
In Fig. \ref{fig:4},
we compare the entity distributions of the best MRR checkpoint of multiple training strategies and find that model performance is related to the cluster formation in the entity distribution.
On the WN18RR dataset, NonNeg and ALLNeg have a higher proportion of entities that are evenly distributed, indicating that they cannot effectively distinguish these entities.
On CoDEx-M and FB15k237, NonNeg has fewer clusters than the other strategies, which explains why it gets relatively weak performance.
On the contrary, there are more clusters and clearer boundaries between different clusters in HaLE's entity distributions, indicating that HaLE can better separate entity embedding vectors.

\end{document}